\documentclass[a4paper,conference]{IEEEtran}

\IEEEoverridecommandlockouts
\usepackage{cite}
\usepackage{amsmath,amssymb,amsfonts}
\usepackage{algorithmic}
\usepackage{graphicx}
\usepackage{textcomp}
\usepackage{url}
\usepackage{hyperref}
\usepackage{xcolor}
\def\BibTeX{{\rm B\kern-.05em{\sc i\kern-.025em b}\kern-.08em
    T\kern-.1667em\lower.7ex\hbox{E}\kern-.125emX}}
\usepackage[spanish]{babel}
\usepackage{subcaption}
\usepackage{balance}
\usepackage[utf8]{inputenc}
\usepackage{newunicodechar}
\newunicodechar{∣}{\mid}

\begin{document}

\title{\textit{On the Shape of Latent Variables in a Denoising VAE-MoG: A Posterior Sampling-Based Study}\\}
\author{
    \IEEEauthorblockN{Fernanda Zapata Bascuñán}
    \IEEEauthorblockA{
        Universidad Nacional de San Martín. \\
        fzapatabascuna@unsam.edu.ar
    }
}

\maketitle
\begin{abstract}
In this work, we explore the latent space of a denoising variational autoencoder with a mixture-of-Gaussians prior (VAE-MoG), trained on gravitational wave data from event GW150914. To evaluate how well the model captures the underlying structure, we use Hamiltonian Monte Carlo (HMC) to draw posterior samples conditioned on clean inputs, and compare them to the encoder's outputs from noisy data. Although the model reconstructs signals accurately, statistical comparisons reveal a clear mismatch in the latent space. This shows that strong denoising performance doesn't necessarily mean the latent representations are reliable—highlighting the importance of using posterior-based validation when evaluating generative models.
\begin{IEEEkeywords}
\textit{Denoising variational autoencoder, latent space analysis, Hamiltonian Monte Carlo, distribution fitting, gravitational wave data.}
\end{IEEEkeywords}

\end{abstract}

\section{Introduction}

The detection and analysis of gravitational waves have significantly advanced in recent years, prompting the development of robust signal processing techniques to enhance event identification and data quality. A major challenge in this domain is the presence of noise from diverse sources, including seismic activity, magnetic field variations, acoustic interference, and electromagnetic disturbances caused by atmospheric or solar events \cite{b1,b2}. These noise sources can mask weak astrophysical signals, making effective denoising a critical task.

Autoencoders have proven effective for noise reduction in signal processing applications \cite{denoising}. Denoising autoencoders (DAEs), in particular, not only improve signal quality but also learn compact and meaningful representations. Variational autoencoders (VAEs) extend this framework by introducing a probabilistic latent space, enabling generative capabilities and uncertainty estimation.

In this work, we implement a denoising variational autoencoder with a mixture-of-Gaussians prior (VAE-MoG), trained on a dataset derived from the GW150914 gravitational wave event. The model uses convolutional layers for feature extraction and incorporates a training objective that combines reconstruction loss with a Kullback-Leibler (KL) divergence term to regularize the latent space.

Beyond denoising performance, our focus is on characterizing the structure of the learned latent representations. To this end, we employ Hamiltonian Monte Carlo (HMC), a gradient-based sampling method from the Markov Chain Monte Carlo (MCMC) family \cite{hmc2}. Unlike standard sampling techniques, HMC efficiently explores complex posterior distributions in high-dimensional spaces by leveraging gradient information to propose informed transitions.

Using HMC, we generate posterior samples over the latent variables and analyze their statistical properties. This empirical approach allows us to investigate the geometry of the latent space without relying on prior assumptions about its distribution, offering insights into the internal consistency and generative behavior of the learned model.

The remainder of this paper is organized as follows: Section II describes the dataset, model architecture, and sampling strategy. Section III outlines the experimental procedure, the underlying hypotheses, and the implementation of the Hamiltonian algorithm. Finally, Section IV concludes the paper and discusses directions for future research.

\section{Previous Work}
An autoencoder consists of two components: an encoder that maps input data to a lower-dimensional latent space, and a decoder that reconstructs the original data from this representation. The model is trained to minimize a reconstruction loss, encouraging the latent space to retain essential information from the input.

There are various implementation methods and internal architectures for autoencoders, such as \textit{Dense Neural Networks (DNNs)}, \textit{Recurrent Neural Networks (RNNs)}, LSTMs, GRUs, Convolutional Layers, and \textit{Spiking Neural Networks (SNNs)}. The decision was made to work with the first architecture, \textit{CNNs}, as the data often does not exhibit explicit temporal dependencies, nor is there a sequential structure observed in the underlying patterns. Therefore, for autoencoders, as in this approach, \textit{CNNs} prove to be a suitable and advantageous choice \cite{deeplearningbook}.
\section{Methods}

Under the assumption of stationary, Gaussian noise uncorrelated \cite{noiseligo,noiseligo2} in each detector as treated in present references of LIGO, we proceeded to create a bank of templates using phenomenological models of gravitational waves through the \textit{PyCBC} \cite{pycbc} and \textit{LALsuite} libraries. The models used are crucial in describing the signals emitted by astrophysical systems, specifically binary black holes (BBH) \cite{templates}, in collision and merger processes. Among the models used in the \textit{dataset} are \textit{IMRPhenom} \cite{imrp}, \textit{SEOBNRv4}, and \textit{Taylorv4} [6,7], which offer theoretical representations of the waveforms that emerge from such astrophysical events and, unlike others, consider features similar to GW150914, such as the angular momentum of black holes.
The dataset was generated by varying only the black hole masses, while keeping other parameters (e.g., angular momentum, inclination, minimum frequency) fixed. This allowed for the creation of templates that capture the full deformation profile of the event.
For this, we chose a maximum spin value for each black hole, 0.7 and 0.9 respectively, the appearance on the sky according to \cite{b1} \cite{noiseligo}, and the mass variation of the black holes with a 0.5 solar masses step between the previously described minima and maxima. Fig. 5 shows a table representing the signals obtained for the set of templates.

\begin{figure}[htbp]
\centering
\includegraphics[scale=0.35]{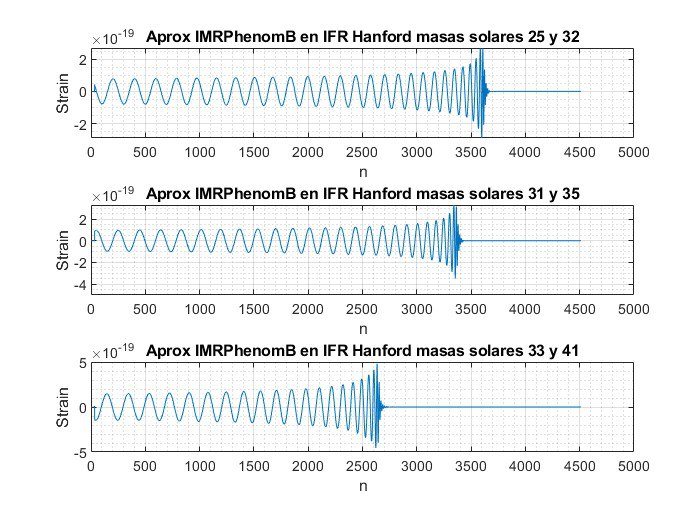}
\caption{Template bank}
\label{fig}
\end{figure}
After generating the synthetic signal set, the responses of three gravitational wave detectors—Hanford (H1), Livingston (L1), and Virgo (V1)—were simulated. Due to differences in their geographic locations and instrumental characteristics, each detector records a slightly different deformation for the same astrophysical event. Our dataset reflects this variability by simulating the response of each analytical signal across all three detectors.

This process resulted in a dataset of 8784 signals, each with a length of 4392 samples. To ensure consistent input size for the model, all signals were normalized by zero-padding shorter signals to match the length of the longest one. This padding preserves the initial timing of each signal and is known to be effective for convolutional neural network architectures \cite{padding}. Notably, architectures that depend heavily on temporal sequence modeling can suffer from degraded training when signals have inconsistent lengths, further justifying this approach.
The inclusion of simulated signals from Hanford, Livingston, and Virgo—despite Virgo not being involved in the original GW150914 event—introduces variability essential for robust training~\cite{pycbc}. Figure 1 illustrates a sample of the dataset, including signals from one of the three detectors and variations in black hole masses generated through an analytical method.

For the noise component, real data from the GW150914 event were obtained from the LIGO archives~\cite{templates, noiseligo}. Using the \textit{PyCBC} library, we computed the power spectral density (PSD) around the event, applying an appropriate cutoff frequency. The final dataset contains noise from both the Hanford (H1) and Livingston (L1) detectors.

\begin{figure}[htbp]
\centering
\includegraphics[scale=0.5]{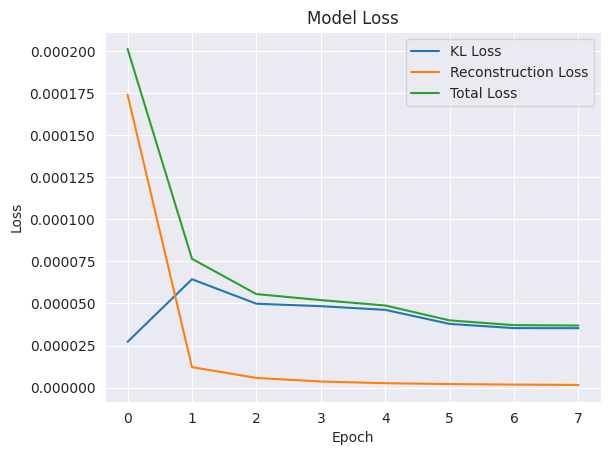}
\caption{Loss function, KL loss and total training loss of the denoising Convolutional VAE training.}
\label{fig}
\end{figure}

\subsection{Procedure}

The Variational Autoencoder (VAE) consists of an encoder and a decoder built with deep neural networks, incorporating a sampling layer that generates latent variables from a parameterized distribution. Our implementation extends the classical VAE architecture by adopting a Mixture of Gaussians (MoG) as the prior distribution over the latent space. This allows the model to capture more flexible and multimodal latent distributions, overcoming limitations of the standard isotropic Gaussian prior \( \mathcal{N}(0, I) \).

The encoder includes several layers: a 1D convolutional layer for feature extraction, followed by max-pooling, dropout, and fully connected (dense) layers that reduce the representation to a latent space of dimension 256. It outputs two vectors, the mean (\( \mathbf{z}_{\text{mean}} \)) and the log-variance (\( \mathbf{z}_{\text{log\_var}} \)), from which the latent variable \( \mathbf{z} \) is sampled via the reparameterization trick. The decoder reconstructs the original input from \( \mathbf{z} \), first mapping the latent vector to a higher-dimensional tensor using dense and reshape layers, then applying a sequence of transposed convolutional layers to generate the denoised output signal. Training is performed using a custom loss function combining mean squared error reconstruction loss and a Kullback-Leibler (KL) divergence term. An adaptive beta schedule gradually increases the weight of the KL divergence during training, encouraging the model to progressively enforce latent space regularization. It was implemented in order to mitigate KL vanishing \cite{klvanishing}. 
The model was trained for 64 epochs with a batch size of 32 using the SGD optimizer with a learning rate of 0.001 and gradient clipping set to 1.0.
\begin{figure}[htbp]
\includegraphics[width=0.475\textwidth]{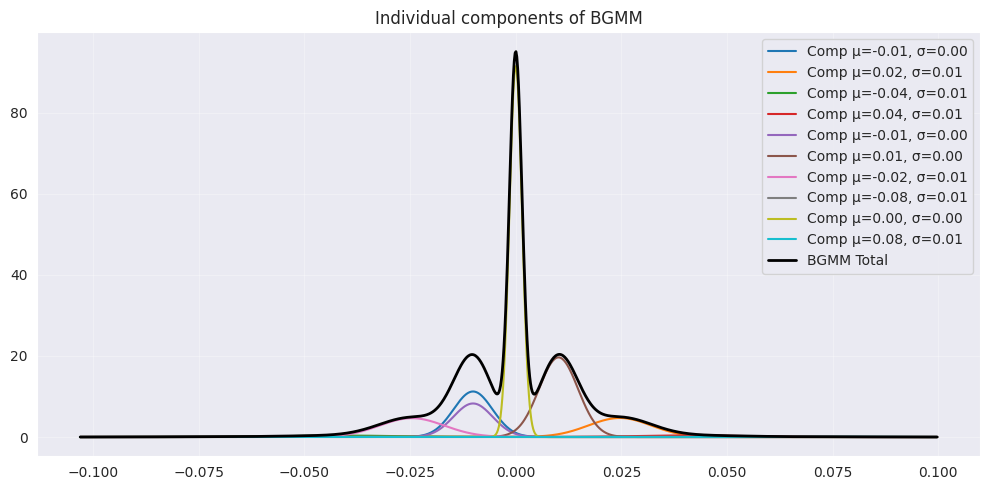}
\caption{Bayesian Gaussian Mixture Model fitted to \( y_{\text{train}} \). The individual Gaussian components are shown, along with the overall estimated density in orange.}
\label{fig}
\end{figure}
\begin{figure}[htbp]
\includegraphics[width=0.49\textwidth]{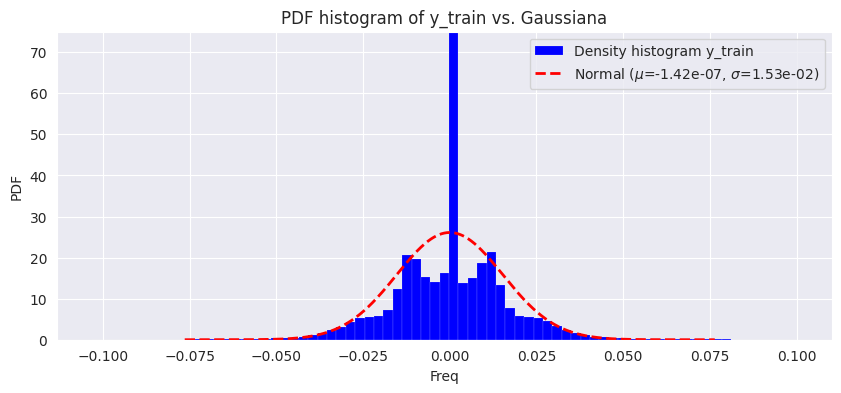}
\caption{ Histogram-based probability density function of the clean training targets \( y_{\text{train}} \), serving as a visual reference of the empirical distribution.}
\label{fig}
\end{figure}

\balance{\section{Posterior Sampling and Evaluation Framework}

\subsection*{Hamiltonian Monte Carlo Sampling of the Latent Distribution}

To approximate the expected data distribution in the latent space, we fitted a parametric Bayesian Gaussian Mixture Model (BGMM) using kernel density estimation (KDE) to the target \( y_{\text{train}} \)~\cite{kde}.The fitted ESBD model of the data is shown in Figure 3, while for visual comparison, the histogram-based probability density function (PDF) of the expected data is presented in Figure 4. The BGMM was limited to a maximum of 10 components to ensure an accurate fit.\\
To perform efficient sampling from the complex posterior distribution induced by this BGMM prior, we implemented a Hamiltonian Monte Carlo (HMC) sampler, following the methodology outlined in~\cite{mc} \cite{hmc}.
The initial points were obtained from the latent space samples $z_{\text{encoder}}$. The HMC sampler employs a leapfrog integrator to propose new states $(q, p)$ by simulating Hamiltonian dynamics defined by the target distribution's log-density and its gradient.

Given the target log-density function and its numerical gradient (estimated via finite differences from the BGMM log-probabilities), the sampler performs:

\begin{itemize}
    \item Initialization of position and momentum variables.
    \item Leapfrog integration steps to propose new samples.
    \item Metropolis acceptance criterion based on Hamiltonian energy.
\end{itemize}

\subsection{Evaluation Hypothesis}
To evaluate the encoder’s denoising capability in latent space, we formulated the following hypothesis: latent codes obtained from noisy inputs (\(z_{\text{noisy}}\)) should approximate the distribution of those derived from clean data (\(z_{\text{clean}}\)). To test this, we:
\begin{itemize}
    \item Fitted a BGMM to \(z_{\text{clean}}\), representing the reference posterior.
    \item Sampled from the BGMM via HMC to obtain high-quality \(z_{\text{clean}}\) samples.
    \item Compared each latent dimension \(d \in \{1, ..., D\}\) using two-sample Kolmogorov–Smirnov (KS) tests.
    \item We assumes weak inter-dimensional dependencies, allowing marginal comparisons as a proxy for joint alignment. 
\end{itemize}
Since marginal KS tests assume weak dependence between dimensions, we assessed pairwise correlations to verify this condition. As shown in Figure 5, the correlations are low, supporting the reliability of the marginal testing strategy.

\begin{figure}[htbp]
\centering
\includegraphics[width=0.49\textwidth]{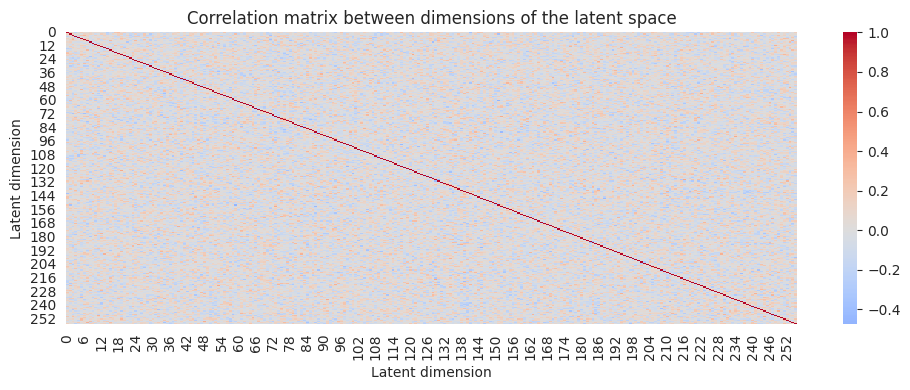}
\caption{Pearson correlation matrix between latent dimensions z extracted by the encoder. Most correlations are close to zero, suggesting relative independence between dimensions.}
\label{fig}
\end{figure}

\section{Results}

\begin{figure}[htbp]
\includegraphics[width=0.5\textwidth]{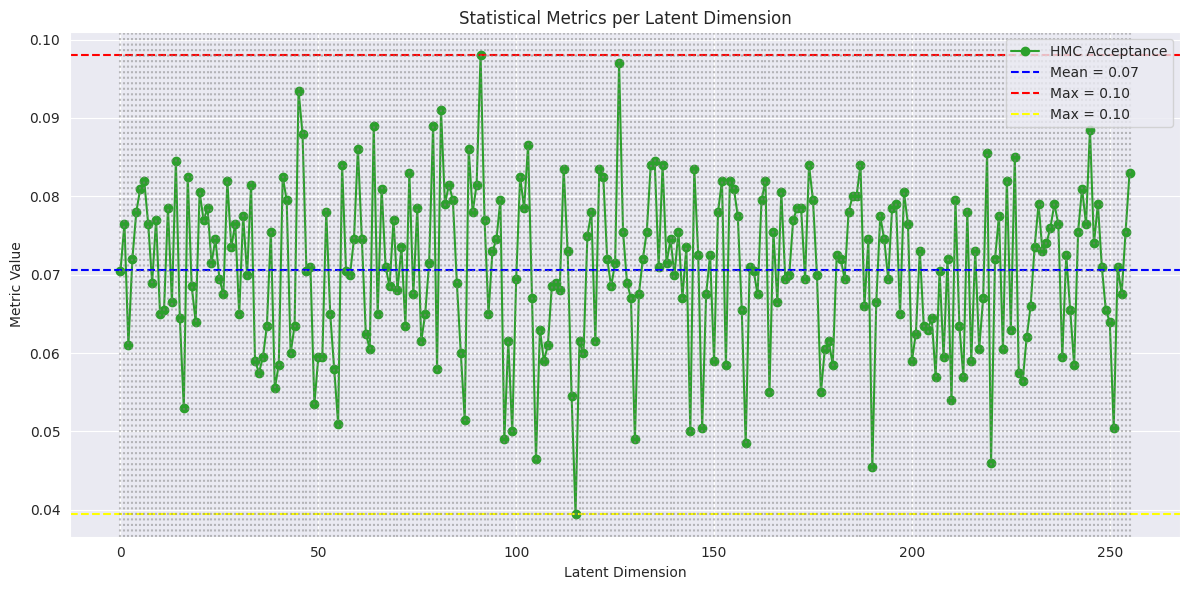}
\caption{Acceptance rates per latent dimension during HMC sampling, reflecting the proportion of Hamiltonian proposals accepted under the Metropolis criterion.}
\label{fig}
\end{figure}

\subsection{Hamiltonian Montecarlo acceptance rate}
The acceptance rate of the HMC sampler varied across latent dimensions, as illustrated in Figure 6. While relatively low, these rates are expected given the complexity of the posterior landscape modeled by the BGMM. Importantly, they remain substantially higher than those observed with alternative target distributions such as univariate Gaussian, Laplace, and spike-and-slab. 
\subsection{Kolmogorov-Smirnov Test on Latent Distributions}
\begin{figure}[htbp]
\includegraphics[width=0.5\textwidth]{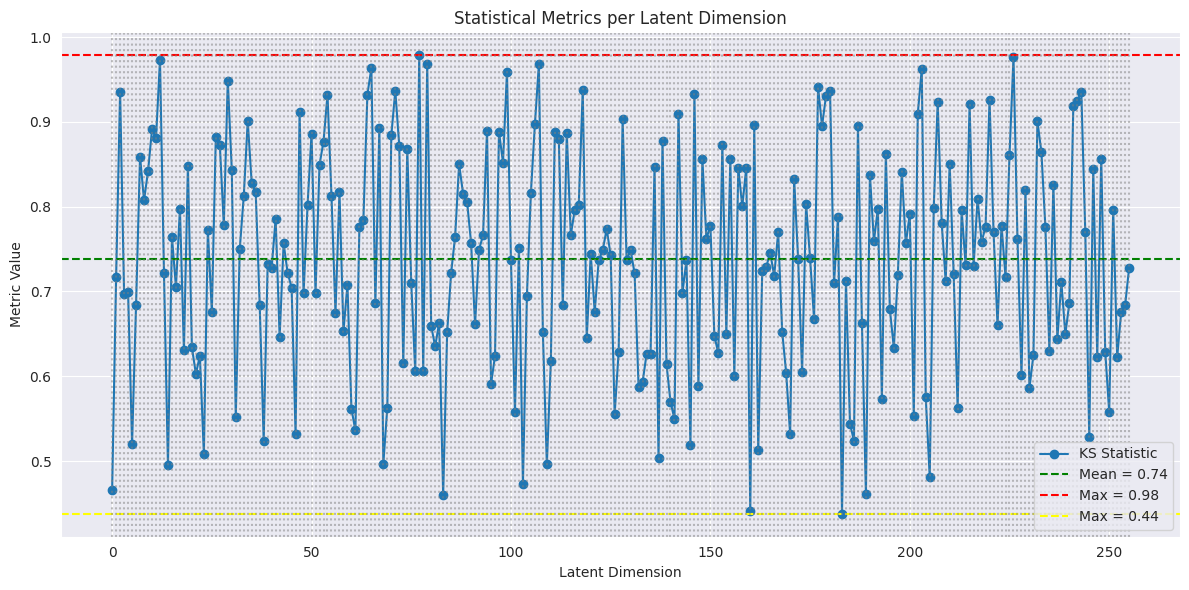}
\caption{ Kolmogorov–Smirnov (KS) statistics across all 256 latent dimensions, shown a significant mismatch between the encoder’s latent representations and posterior samples obtained via HMC.}
\label{fig}
\end{figure}
To evaluate the similarity between latent representations obtained from noisy inputs and those sampled from the BGMM posterior via Hamiltonian Monte Carlo (HMC), we performed two-sample Kolmogorov–Smirnov (KS) tests across all 256 latent dimensions. The consistently high KS statistics indicate significant differences between the two distributions, as illustrated in Figure 7. These findings reveal a systematic discrepancy between the encoder’s outputs and the true posterior, suggesting that despite visually accurate reconstructions, the latent space structure is not preserved in the presence of noise.

\section{Conclusions}
This work investigates the latent space geometry of a denoising variational autoencoder (DVAE) with a mixture-of-Gaussians prior, trained on noisy gravitational wave signals. By leveraging Hamiltonian Monte Carlo (HMC) sampling from the posterior conditioned on clean data, we establish a principled method to probe the structure and fidelity of the learned latent distribution.

Through HMC, we obtain high-quality samples that reflect the underlying generative geometry, enabling a detailed comparison with the encoder outputs from corrupted inputs. This comparison, assessed statistically across latent dimensions, reveals a persistent mismatch—suggesting that the encoder does not recover the full posterior structure, even when reconstructions appear accurate.

These results emphasize the utility of posterior sampling as a diagnostic tool for latent consistency. Rather than relying solely on reconstruction loss, our approach offers a deeper lens into how well the inference mechanism aligns with the generative process—an essential consideration for models applied to scientific domains where interpretability and structure matter.

\subsection{Future Work}
Future work should improve latent space modeling and encoder robustness, as the KS test reveals the encoder’s difficulty recovering correct latent distributions under noise. The current BGMM prior also fails to capture key structural features, affecting posterior accuracy. To address this, we propose exploring flexible priors (e.g., normalizing flows), applying latent regularization techniques like MMD or adversarial penalties, pruning unstable dimensions, and refining the decoder architecture. These steps aim to enhance latent fidelity and downstream task performance in noisy data.

\vspace{12pt}
}

\end{document}